\documentclass[twocolumn,letterpaper]{IEEEAerospaceCLS}  %

\usepackage[]{graphicx}    %
\usepackage{svg}
\usepackage[colorlinks=true, urlcolor=blue, linkcolor=black, citecolor=black]{hyperref} %
\newcommand{\ignore}[1]{}  %
\usepackage{booktabs}
\renewcommand{\toprule}{\specialrule{.08em}{0pt}{.65ex}}
\renewcommand{\midrule}{\specialrule{.05em}{.4ex}{.65ex}}
\renewcommand{\bottomrule}{\specialrule{.08em}{.4ex}{0pt}}

\usepackage[caption=false,font=footnotesize]{subfig}

\begin{document}
\title{Embedded Bare-Metal Radar-Inertial Odometry}

\author{%
Nicolai Adil {\O}yen Aatif \qquad Morten Nissov \qquad Kostas Alexis \qquad \\
Department of Engineering Cybernetics\\
Norwegian University of Science and Technology\\
O. S. Bragstads Plass 2D, 7034 Trondheim, Norway\\
nicolaiadil.aatif@gmail.com \quad morten.c.nissov@gmail.com \quad konstantinos.alexis@ntnu.no
\thanks{\footnotesize 979-8-3195-1048-8/27/\$31.00 \copyright2027 IEEE}
}

\maketitle

\thispagestyle{plain}
\pagestyle{plain}

\begin{abstract}
Compact extraplanetary rovers and micro aerial vehicles require robust state estimation frameworks designed to operate under strict computational constraints in unforgiving environments. Typical solutions involving vision- or LiDAR-based sensing are computationally expensive and vulnerable to environments with perceptual degradation, making them poorly suited for resource-constrained platforms and austere conditions. Alternatively, Frequency Modulated Continuous Wave (FMCW) radar offers both robustness and computational efficiency by directly providing velocity measurements coupled with inherent resilience to perceptual degradation. These factors enable robust estimation, thereby reducing reliance on human operators for supervision to ensure platform safety in challenging environments. In this manuscript, we propose an embedded radar-inertial estimator tailored for low-compute platforms. All sensor drivers, data processing, and aided inertial navigation are performed on a single-core microcontroller, demonstrating its computational efficiency. 
Flight experiments show translational APE of $0.51-0.82\,\mathrm{m}$ and RPE of $<$3\,\% against motion capture, alongside closed-loop flight through an unmodified PX4 stack. %
The firmware and printed circuit board design are openly available on GitHub at \href{https://github.com/ntnu-arl/embedded_rio}{ntnu-arl/embedded\_rio} and \href{https://github.com/ntnu-arl/embedded_rio-pcb}{ntnu-arl/embedded\_rio-pcb} respectively.
\end{abstract}

\tableofcontents

\section{Introduction}

\subsection{Motivation}

Reliable state estimation is required for autonomous operation of small robotic platforms. For micro aerial vehicles and planetary rovers, this estimation must often be performed onboard under strict size, weight, power, and cost (SWaP-C) constraints, together with limited memory and processing capability. These constraints make the computational cost of the estimation pipeline an important design consideration.

Visual-inertial and LiDAR-inertial methods can provide accurate motion estimates, but they typically require a substantial perception front-end. Image-based methods often depend on feature extraction and tracking, while LiDAR-based methods commonly rely on scan registration \cite{survey_vslam}. In addition to the associated computational requirements, both sensing modalities may degrade in challenging environments. Vision is sensitive to illumination and texture, while LiDAR measurements can be affected by geometric self-similarity alongside airborne particles such as dust, fog, and smoke \cite{survey_radars}.

FMCW radar provides a different and complementary sensing modality. Modern radar devices perform range, Doppler, and angle-of-arrival processing internally and output a sparse set of detections containing both position and radial-speed information. The host processor can therefore use the radar measurements for inertial aiding without performing image processing, scan registration, or data association between successive radar frames. Together with the small size of modern antenna-on-package radar devices \cite{IWR6843AOPEVM}, this makes radar particularly suitable for state estimation on platforms with limited computational resources.

\subsection{Related Work}

Radar-inertial estimators differ primarily in how Doppler measurements are incorporated into the estimator. One family estimates a three-dimensional velocity from the radial-speed returns of an entire radar frame using formulations such as \cite{kellner_ego_motion}, and uses this velocity as an aiding measurement in an EKF \cite{doer2020ekfrio}. Extensions of this approach include online estimation of the temporal offset between the radar and IMU \cite{ekf_rio_temporal_cal}. Estimating a complete velocity vector from a radar frame requires sufficient geometric diversity among the detections, which may not be available with a restricted field of view or sparse returns.

An alternative is to incorporate each radar detection individually as a scalar radial-speed measurement along its bearing direction \cite{nissov_robusthighspeed,nissov2024degradationresilientlidarradarinertialodometry,michalczyk2026constrained}. This avoids the need for an intermediate three-dimensional velocity estimate from each frame. Other approaches do not use Doppler information and instead associate radar landmarks between consecutive frames \cite{radar_only_ego_motion}, introducing an explicit data-association stage.

With radial-speed aiding alone, the vertical position and yaw remain unobservable and will accumulate drift over longer trajectories. Heading drift has been reduced using structural assumptions on the surrounding environment \cite{radar_inertial_dataset}, while vertical drift can be limited using barometric aiding \cite{doer2020ekfrio}. Additional exteroceptive sensors have also been introduced, including LiDAR \cite{nissov2024degradationresilientlidarradarinertialodometry} and vision \cite{nissov2026rvio}, at the cost of additional sensing and computation.

Computationally constrained radar-inertial navigation has been demonstrated on small aerial platforms \cite{michalczyk2023closedloop}. These systems nevertheless employ powerful companion computers running general-purpose, non-real-time operating systems. The present work retains per-detection radial-speed aiding together with barometric altitude aiding, while moving the complete estimation pipeline to a bare-metal, single-core microcontroller.

\subsection{Contributions}

The contributions of this work are as follows.

\begin{itemize}
\item[$\bullet$] A tightly coupled radar-inertial error-state Kalman filter using per-detection radial-speed aiding, anchored differential barometric aiding, second-order measurement underweighting, and online estimation of the radar--IMU extrinsics. The estimator executes in single precision (IEEE 754) on the single core of a $600\,\mathrm{MHz}$ Cortex-M7 microcontroller, together with the sensor drivers and measurement handling, without an operating system, dynamic memory allocation, or hardware acceleration.

\item[$\bullet$] A flight-ready sensor module integrating the radar, IMU, barometer, and microcontroller on a compact assembly sized for integration on a small multirotor.

\item[$\bullet$] A characterization of the computational and memory footprint of the embedded implementation, together with a parity analysis against a desktop replay compiled from the same estimator source.

\item[$\bullet$] Flight validation in open-field, forest, and motion-capture environments, including pose-error evaluation against motion-capture ground truth and closed-loop position control through an unmodified PX4 flight stack using its documented external-odometry interface.
\end{itemize}

The remainder of this manuscript is organized as follows. Section~\ref{sec:system_architecture} describes the system architecture and the partitioning of computation between the radar, microcontroller, and autopilot. Section~\ref{sec:estimator} presents the radar-inertial estimator and its measurement models. Section~\ref{sec:embedded} describes the embedded implementation and flight-stack interface. Sections~\ref{sec:experimental_setup} and \ref{sec:results} present the experimental setup and results, respectively, and Section~\ref{sec:conclusion} concludes the paper.

\section{System Architecture}
\label{sec:system_architecture}

\begin{figure}[t]
    \centering
    \includegraphics[width=\columnwidth]
    {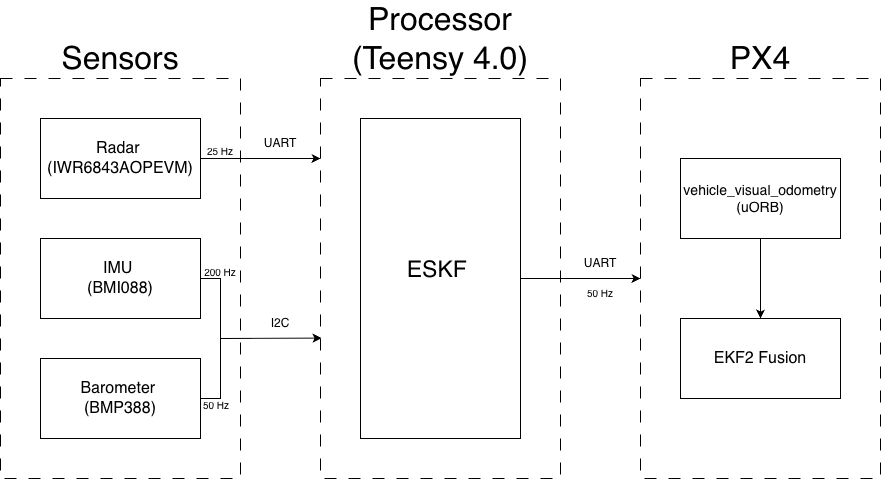}
    \caption{System architecture and division of computation between the radar SoC, microcontroller (Teensy 4.0), and flight controller.}
    \label{fig:system_overview}
\end{figure}

The system is organised such that radar signal processing is performed on the radar system-on-chip (SoC), while sensor handling and radar-inertial state estimation are performed on a single microcontroller. The resulting odometry is subsequently provided to the flight controller through its standard external-odometry interface. Figure~\ref{fig:system_overview} shows the resulting data flow.

\subsection{Sensor Suite}
\label{sec:sensor_suite}

\begin{figure}[t]
    \centering
    \includegraphics[
        width=\columnwidth,
        trim=0 15cm 0 15cm,
        clip
    ]{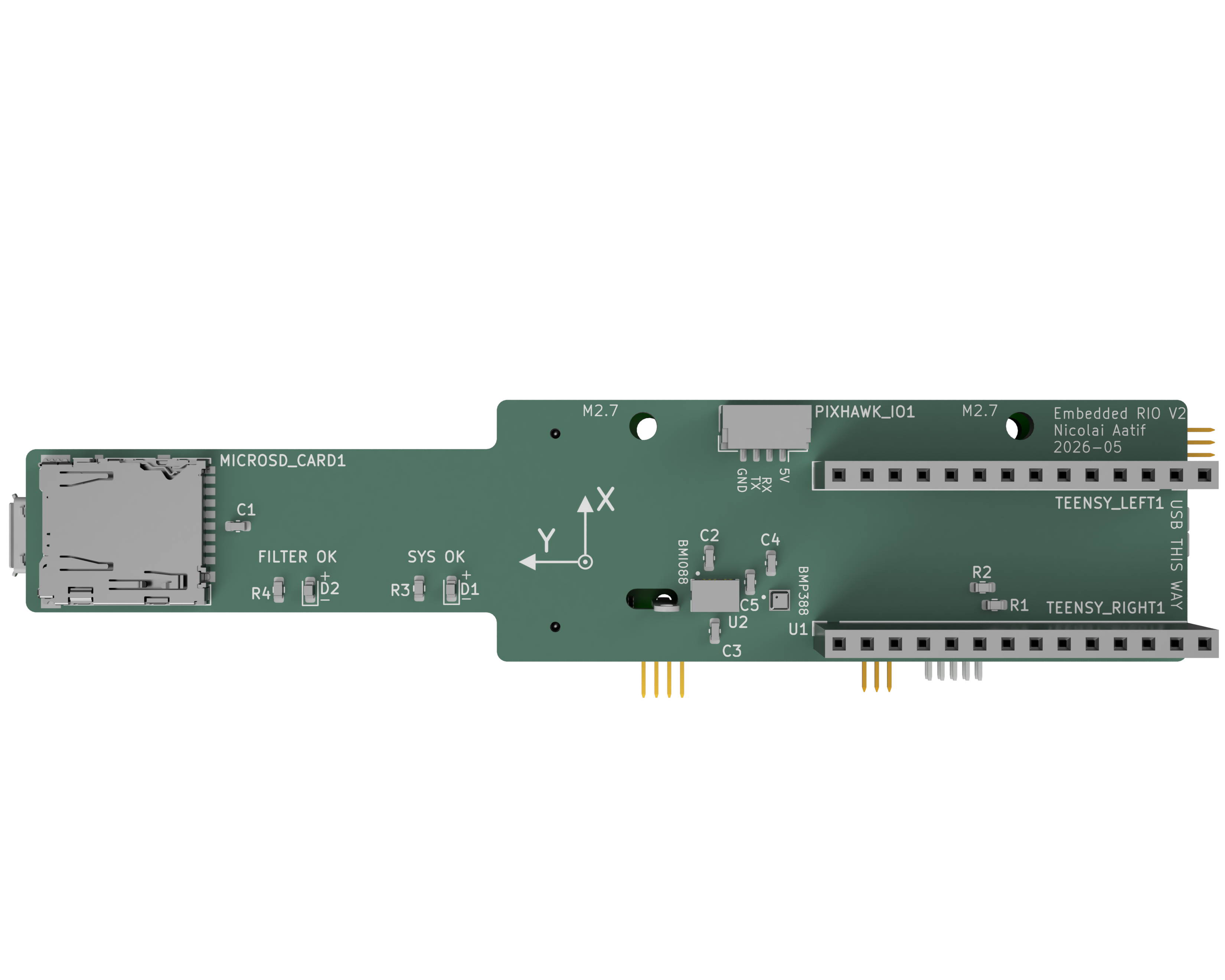}
    \par\vspace{-1.5em}
    {\small (a)}
    \vspace{0.5em}
    \includegraphics[
        width=\columnwidth,
        trim=0 15cm 0 15cm,
        clip
    ]{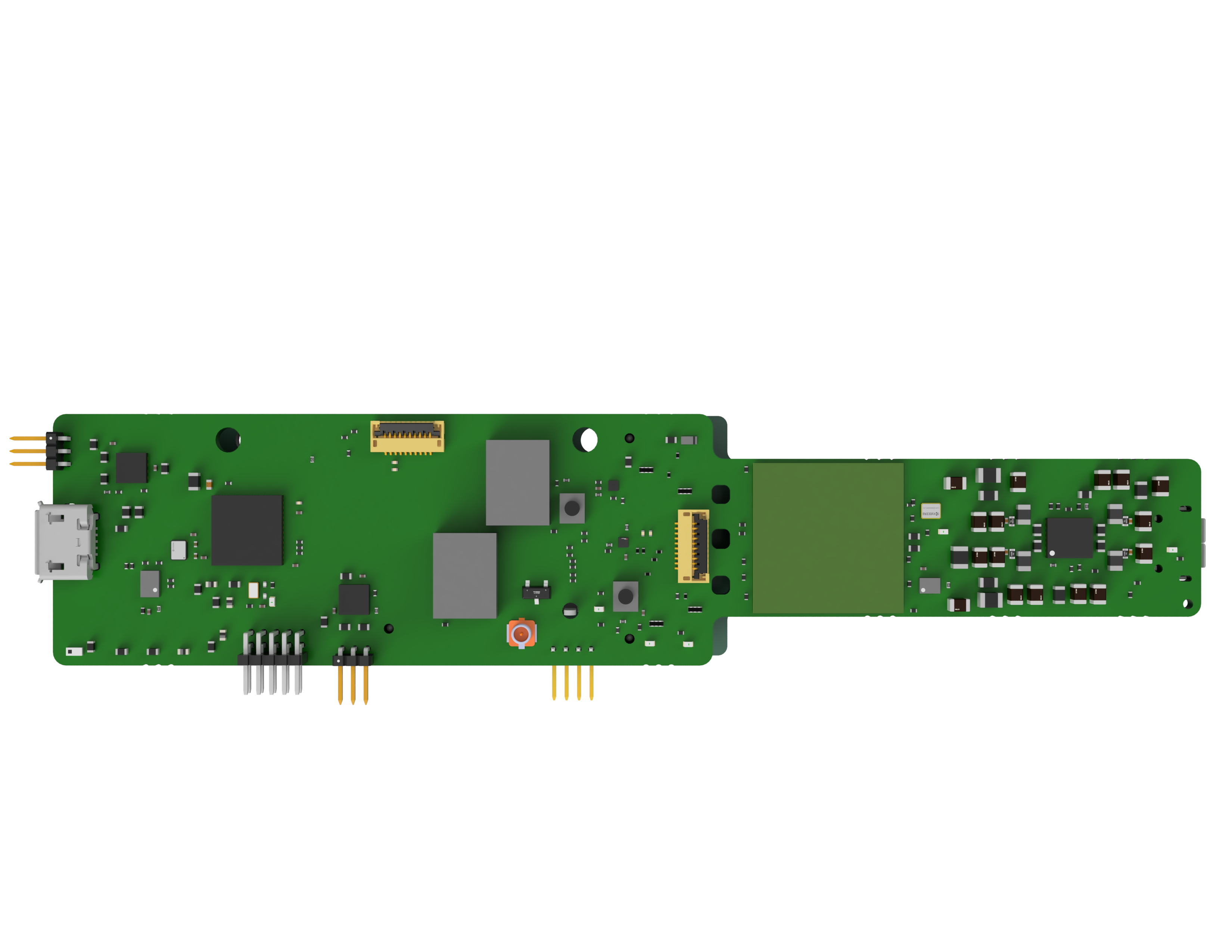}
    \par\vspace{-1.5em}
    {\small (b)}
    \caption{Custom radar-inertial carrier board with a $115 \times 25\,\mathrm{mm}$ footprint
    (a) Top view showing the custom PCB. (b) Bottom view showing the radar.}
    \label{fig:carrier_board}
\end{figure}

The sensing and estimation hardware is integrated on a custom carrier board built around a $600\,\mathrm{MHz}$ Cortex-M7 microcontroller, shown in Figure~\ref{fig:carrier_board}. 
The sensor suite consists of a Texas Instruments IWR6843AOPEVM $60\,\mathrm{GHz}$ FMCW radar \cite{IWR6843AOPEVM}, a Bosch BMI088 inertial measurement unit (IMU), and a Bosch BMP388 barometric pressure sensor. The same board interfaces directly with a Pixhawk 6C Mini flight controller.

The radar is mounted at the front of the platform with its azimuth plane approximately parallel to the horizontal plane, placing its largest field of view in the direction of flight. The IMU is located near the geometric centre of the platform, while the barometer is mounted adjacent to it. The radar--IMU translation and rotation are initialised from the mechanical geometry of the assembly and are subsequently refined online as part of the estimator described in Section~\ref{sec:estimator}.

The radar configuration used throughout this work operates at $25\,\mathrm{Hz}$ and provides a maximum unambiguous range of $13.7\,\mathrm{m}$ and radial speed of $9.68\,\mathrm{m/s}$, with range and radial-speed resolutions of $0.21\,\mathrm{m}$ and $0.15\,\mathrm{m/s}$, respectively. Its azimuth and elevation fields of view are both approximately $\pm60^\circ$. The IMU is sampled at $200\,\mathrm{Hz}$, while barometric pressure is sampled at $50\,\mathrm{Hz}$.

\subsection{Compute Partitioning and Timing}
\label{sec:compute_partitioning}

The computational pipeline is divided into three stages. First, the radar SoC performs the range, Doppler, and angle-of-arrival processing of the received FMCW signals. The microcontroller therefore does not process raw radar samples, but receives a sparse list of detections containing Cartesian position and measured radial speed.

Second, the Cortex-M7 microcontroller samples the IMU and barometer, parses the radar detection stream, and executes the complete radar-inertial ESKF. The $200\,\mathrm{Hz}$ IMU measurements drive the inertial prediction, while barometric measurements provide scalar updates at $50\,\mathrm{Hz}$. Radar frames arrive every $40\,\mathrm{ms}$; each detection within a frame is processed individually as a scalar radial-speed update. Consequently, no scan-to-scan radar registration or intermediate three-dimensional radar velocity estimate is performed on the microcontroller.

Finally, the estimated state is transmitted to the Pixhawk over MAVLink using the standard \texttt{ODOMETRY} message \cite{mavlink_developer_guide}. PX4 exposes these messages through the \texttt{vehicle\_visual\_odometry} interface to its internal EKF2 \cite{px4_user_guide}. The Pixhawk is therefore a downstream consumer of the radar-inertial estimate rather than part of the proposed estimator. The estimator itself, including sensor handling, inertial propagation, and all radar and barometric updates, executes on the single microcontroller core.

\section{Radar-Inertial Estimator}
\label{sec:estimator}

Throughout this section, the world, IMU, and radar frames are denoted $\{\mathrm{W}\}$, $\{\mathrm{I}\}$, and $\{\mathrm{R}\}$, respectively. A quantity expressed in frame $\{\mathrm{A}\}$ carries the leading subscript $\mathrm{A}$, while ${}_{\mathrm{A}}\mathbf{R}_{\mathrm{B}}$ rotates a vector from $\{\mathrm{B}\}$ to $\{\mathrm{A}\}$. The operator $[\cdot]_\times$ denotes the skew-symmetric matrix associated with the vector cross product.

\subsection{Frequency-Modulated Continuous-Wave Radar}
\label{sec:fmcw_radar}

The FMCW radar provides the range, radial speed, and angle of arrival of detected reflectors \cite{ti_radar_fundamentals,richards_radar}. Range, Doppler, and angle processing is performed on the radar SoC, and the microcontroller receives a sparse point cloud. For detection $i$, let ${}_{\mathrm{R}}\mathbf{p}_i$ denote its Cartesian position in the radar frame and $\tilde v_{r,i}$ its measured radial speed. The corresponding bearing is
\begin{equation}
    {}_{\mathrm{R}}\boldsymbol{\mu}_i
    =
    \frac{{}_{\mathrm{R}}\mathbf{p}_i}
         {\|{}_{\mathrm{R}}\mathbf{p}_i\|}.
    \label{eq:radar_bearing}
\end{equation}

A reflector that is stationary in $\{\mathrm{W}\}$ has zero world-frame velocity. Its distance from a moving radar nevertheless changes as the radar moves through the environment, producing a non-zero relative range rate and therefore a Doppler shift. Under the static-scene 
assumptions used in \cite{kellner_ego_motion}, the measured radial speed is related to the radar ego-velocity by
\begin{equation}
    v_{r,i}
    =
    -\,{}_{\mathrm{R}}\boldsymbol{\mu}_i^\top
    {}_{\mathrm{R}}\mathbf{v}_{\mathrm{WR}},
    \label{eq:radial_speed}
\end{equation}
where ${}_{\mathrm{R}}\mathbf{v}_{\mathrm{WR}}$ is the velocity of the radar with respect to the world frame, expressed in the radar frame. Each stationary detection therefore provides a scalar measurement of radar ego-motion along its bearing direction, without requiring correspondence between consecutive radar frames.

\subsection{State and Process Model}
\label{sec:state_process}

State estimation is performed using an error-state Kalman filter (ESKF). The nominal state is
\begin{equation}
\mathbf{x}
=
\big(
{}_{\mathrm{W}}\mathbf{p}_{\mathrm{WI}},
{}_{\mathrm{W}}\mathbf{v}_{\mathrm{WI}},
\mathbf{a}_b,
{}_{\mathrm{W}}\mathbf{R}_{\mathrm{I}},
\boldsymbol{\omega}_b,
{}_{\mathrm{I}}\mathbf{p}_{\mathrm{IR}},
{}_{\mathrm{I}}\mathbf{R}_{\mathrm{R}}
\big),
\label{eq:rio_state}
\end{equation}
where
${}_{\mathrm{W}}\mathbf{p}_{\mathrm{WI}}$,
${}_{\mathrm{W}}\mathbf{v}_{\mathrm{WI}}$, and
${}_{\mathrm{W}}\mathbf{R}_{\mathrm{I}}$ denote the IMU position, velocity, and attitude. The accelerometer and gyroscope biases are $\mathbf{a}_b$ and
$\boldsymbol{\omega}_b$ respectively, while
${}_{\mathrm{I}}\mathbf{p}_{\mathrm{IR}}$ and
${}_{\mathrm{I}}\mathbf{R}_{\mathrm{R}}$ describe the radar--IMU
extrinsics. Including the extrinsics gives a 21-dimensional error state and allows their estimates to be refined online.

The nominal navigation state follows the standard inertial kinematics
\cite{sola2017quaternionkinematicserrorstatekalman,sf2024},
\begin{align}
{}_{\mathrm{W}}\dot{\mathbf{p}}_{\mathrm{WI}}
&=
{}_{\mathrm{W}}\mathbf{v}_{\mathrm{WI}},\\
{}_{\mathrm{W}}\dot{\mathbf{v}}_{\mathrm{WI}}
&=
{}_{\mathrm{W}}\mathbf{R}_{\mathrm{I}}
\left(
{}_{\mathrm{I}}\tilde{\mathbf{a}}
-
\mathbf{a}_b
\right)
+
{}_{\mathrm{W}}\mathbf{g},\\
{}_{\mathrm{W}}\dot{\mathbf{R}}_{\mathrm{I}}
&=
{}_{\mathrm{W}}\mathbf{R}_{\mathrm{I}}
\left[
{}_{\mathrm{I}}\tilde{\boldsymbol{\omega}}_{\mathrm{WI}}
-
\boldsymbol{\omega}_b
\right]_\times,\\
{}_{\mathrm{I}}\dot{\mathbf{a}}_b
&=
-p_{ab}\,\mathbf{a}_b,
\\
{}_{\mathrm{I}}\dot{\boldsymbol{\omega}}_b
&=
-p_{\omega b}\,\boldsymbol{\omega}_b ,
\label{eq:rio_process}
\end{align}
where $p_{ab}$ and $p_{\omega b}$ are the inverse time constants of the first-order bias models. Process noise is applied to the inertial measurements and bias dynamics during covariance propagation. The radar--IMU extrinsics are modeled as constant during propagation.

The estimator is initialised while the platform is stationary. Roll and pitch are obtained from the gravity direction, while yaw is initialised to zero and defines the initial world-frame heading. The IMU biases are initialised to zero. The radar extrinsics are initialised from the mechanical geometry described in Section~\ref{sec:sensor_suite}, with non-zero prior covariance to allow subsequent refinement from radar measurements.

\subsection{Radar Radial-Speed Factor}
\label{sec:radar_factor}

The radar and IMU generally do not share the same origin. Platform rotation therefore induces an additional radar velocity through the radar--IMU lever arm. Substituting the rigid-body velocity relation into \eqref{eq:radial_speed} gives the predicted radial-speed measurement
\begin{equation}
\begin{split}
h_i(\mathbf{x})
={}&
-\,{}_{\mathrm{R}}\boldsymbol{\mu}_i^\top
{}_{\mathrm{R}}\mathbf{R}_{\mathrm{I}}
\Big(
{}_{\mathrm{W}}\mathbf{R}_{\mathrm{I}}^\top
{}_{\mathrm{W}}\mathbf{v}_{\mathrm{WI}}
\\
&\quad+
\big(
{}_{\mathrm{I}}\tilde{\boldsymbol{\omega}}_{\mathrm{WI}}
-
\boldsymbol{\omega}_b
\big)
\times
{}_{\mathrm{I}}\mathbf{p}_{\mathrm{IR}}
\Big),
\end{split}
\label{eq:radar_factor}
\end{equation}
where
${}_{\mathrm{R}}\mathbf{R}_{\mathrm{I}}
=
{}_{\mathrm{I}}\mathbf{R}_{\mathrm{R}}^\top$.
The scalar innovation associated with detection $i$ is
\begin{equation}
    e_i
    =
    \tilde v_{r,i}-h_i(\mathbf{x}).
    \label{eq:radar_residual}
\end{equation}

Using the 21-dimensional error state
\begin{equation}
\begin{split}
\delta\mathbf{x}
=
\big(
&{}_{\mathrm{W}}\delta\mathbf{p}_{\mathrm{WI}},
{}_{\mathrm{W}}\delta\mathbf{v}_{\mathrm{WI}},
{}_{\mathrm{I}}\delta\mathbf{a}_b,
\delta\boldsymbol{\theta}_{\mathrm{WI}},\\
&{}_{\mathrm{I}}\delta\boldsymbol{\omega}_b,
{}_{\mathrm{I}}\delta\mathbf{p}_{\mathrm{IR}},
\delta\boldsymbol{\theta}_{\mathrm{IR}}
\big),
\end{split}
\label{eq:error_state}
\end{equation}
linearisation of \eqref{eq:radar_factor} gives five non-zero
measurement-Jacobian blocks. The velocity block is
\begin{equation}
\mathbf{H}_{\delta\mathbf{v}}
=
-\,{}_{\mathrm{R}}\boldsymbol{\mu}_i^\top
{}_{\mathrm{R}}\mathbf{R}_{\mathrm{I}}
{}_{\mathrm{W}}\mathbf{R}_{\mathrm{I}}^\top .
\label{eq:radar_jac_v}
\end{equation}
The Jacobian of the attitude error is
\begin{equation}
\mathbf{H}_{\delta\boldsymbol{\theta}_{\mathrm{WI}}}
=
-\,{}_{\mathrm{R}}\boldsymbol{\mu}_i^\top
{}_{\mathrm{R}}\mathbf{R}_{\mathrm{I}}
\left[
{}_{\mathrm{W}}\mathbf{R}_{\mathrm{I}}^\top
{}_{\mathrm{W}}\mathbf{v}_{\mathrm{WI}}
\right]_\times .
\label{eq:radar_jac_att}
\end{equation}
The gyroscope-bias block becomes
\begin{equation}
\mathbf{H}_{\delta\boldsymbol{\omega}_b}
=
-\,{}_{\mathrm{R}}\boldsymbol{\mu}_i^\top
{}_{\mathrm{R}}\mathbf{R}_{\mathrm{I}}
\left[
{}_{\mathrm{I}}\mathbf{p}_{\mathrm{IR}}
\right]_\times ,
\label{eq:radar_jac_gyro_bias}
\end{equation}
while the radar--IMU translation block is
\begin{equation}
\begin{split}
\mathbf{H}_{\delta\mathbf{p}_{\mathrm{IR}}}
=
-\,{}_{\mathrm{R}}\boldsymbol{\mu}_i^\top
{}_{\mathrm{R}}\mathbf{R}_{\mathrm{I}}
\left[
{}_{\mathrm{I}}\tilde{\boldsymbol{\omega}}_{\mathrm{WI}}
-
\boldsymbol{\omega}_b
\right]_\times .
\end{split}
\label{eq:radar_jac_extrinsic_pos}
\end{equation}
Finally, the radar--IMU attitude block is
\begin{equation}
\begin{split}
\mathbf{H}_{\delta\boldsymbol{\theta}_{\mathrm{IR}}}
={}&
-\,{}_{\mathrm{R}}\boldsymbol{\mu}_i^\top
\Bigg[
{}_{\mathrm{R}}\mathbf{R}_{\mathrm{I}}
\Big(
{}_{\mathrm{W}}\mathbf{R}_{\mathrm{I}}^\top
{}_{\mathrm{W}}\mathbf{v}_{\mathrm{WI}}
\\
&\quad+
\left[
{}_{\mathrm{I}}\tilde{\boldsymbol{\omega}}_{\mathrm{WI}}
-
\boldsymbol{\omega}_b
\right]_\times
{}_{\mathrm{I}}\mathbf{p}_{\mathrm{IR}}
\Big)
\Bigg]_\times .
\end{split}
\label{eq:radar_jac_extrinsic_att}
\end{equation}
The position and accelerometer-bias blocks are zero. Collecting all
blocks gives
\begin{equation}
\begin{split}
\mathbf{H}_i
=
\big[
&\mathbf{0}_{1\times3},\
\mathbf{H}_{\delta\mathbf{v}},\
\mathbf{0}_{1\times3},\
\mathbf{H}_{\delta\boldsymbol{\theta}_{\mathrm{WI}}},\\
&\mathbf{H}_{\delta\boldsymbol{\omega}_b},\
\mathbf{H}_{\delta\mathbf{p}_{\mathrm{IR}}},\
\mathbf{H}_{\delta\boldsymbol{\theta}_{\mathrm{IR}}}
\big]
\end{split}
\label{eq:radar_jacobian}
\end{equation}

The expressions show that a radar detection aids velocity and also provides information about platform attitude, gyroscope bias, and the radar--IMU extrinsics through their coupling in the radial-speed model. In particular, sensitivity to gyroscope bias is proportional to the radar--IMU lever arm and is therefore weak when the two sensor origins are close.

Each detection is processed as an independent scalar update, with the posterior after one accepted detection forming the prior for the next. Measurements inconsistent with the static-scene model, including returns from moving objects, multipath, and self-reflections, are rejected using a chi-squared innovation gate. Before applying an update, the normalised innovation squared $\gamma_i$ is computed from the innovation $e_i$ and its variance $S_i$ as
\begin{equation}
    \gamma_i = \frac{e_i^2}{S_i}.
    \label{eq:radar_gate}
\end{equation}
For a chosen gating threshold $\sigma_{\mathrm{gate}}$ in units of standard deviations, a measurement is accepted if $\gamma_i \le \sigma_{\mathrm{gate}}^2$ \cite{sf2024}. Rejected detections do not affect the state or covariance.

The radar measurement model is nonlinear in all of the estimated quantities. To account for second-order measurement curvature, the innovation variance is augmented using the underweighting formulation of \cite{zanetti2010underweighting,nasa_nav_filter_best_practices},
\begin{align}
    S_i
    &=
    \mathbf{H}_i\mathbf{P}\mathbf{H}_i^\top
    +
    \sigma_{v_r}^{2}
    +
    B_i ,\\
    B_i
    &=
    \frac{1}{2}
    \operatorname{tr}
    \left(
        \mathbf{H}_{xx,i}\mathbf{P}
        \mathbf{H}_{xx,i}\mathbf{P}
    \right),
    \label{eq:radar_innovation}
\end{align}
where $\mathbf{P}$ is the prior error-state covariance, and $\sigma_{v_r}$ is the radial-speed measurement standard deviation set according to the chirp configuration following \cite{nissov_characterization}, assuming no uncertainty in the point bearing. Furthermore, $\mathbf{H}_{xx,i}$ is the Hessian of the scalar measurement model with respect to the error state. The Hessian is obtained by central finite differences of the analytical Jacobian and is symmetrised as
\begin{equation}
    \mathbf{H}_{xx,i}
    \leftarrow
    \frac{1}{2}
    \left(
        \mathbf{H}_{xx,i}
        +
        \mathbf{H}_{xx,i}^{\top}
    \right).
    \label{eq:hessian_sym}
\end{equation}
Numerical round-off can produce a small negative value of $B_i$, in which case it is clipped to zero such that the correction can only increase the innovation variance.

\subsection{Anchored Differential Barometric Factor}
\label{sec:baro_factor}

Radial-speed aiding does not provide observability over absolute position, so drift accumulates in all position components, with vertical position being particularly susceptible to persistent errors in the estimated vertical velocity.%
Barometric pressure is therefore incorporated as an additional scalar aiding measurement using the anchored differential formulation in \cite{doer2020ekfrio,girod2024brio}. Let $h(P)$ denote the pressure-to-altitude mapping of the standard atmosphere model \cite{nasa_glenn_eam}. At the first valid pressure measurement $P_0$, the corresponding barometric altitude and filter height are stored. Subsequent pressure measurements give
\begin{equation}
    \Delta\tilde z_i
    =
    h(P_i)-h(P_0),
    \label{eq:baro_delta}
\end{equation}
and the corresponding innovation is
\begin{equation}
    e_{z,i}
    =
    \Delta\tilde z_i
    -
    \left(
        {}_{\mathrm{W}}p_{\mathrm{WI},z}
        -
        z_{\mathrm{state},0}
    \right).
    \label{eq:baro_residual}
\end{equation}
The measurement is linear in vertical position, giving
\begin{equation}
    \mathbf{H}_{z}
    =
    \left[
    \mathbf{0}_{1\times2},\
    1,\
    \mathbf{0}_{1\times18}
    \right].
    \label{eq:baro_jacobian}
\end{equation}

Differencing against the initial measurement removes constant terms common to both pressure readings, including the absolute pressure offset, the constant component of the atmospheric-model error, and the fixed vertical displacement between the barometer and IMU. The measurement standard deviation is set to $\sigma_{\Delta z}=0.20\,\mathrm{m}$ to account for pressure disturbances beyond the sensor noise floor. The barometric update thereby aids vertical position and limits vertical drift.

\section{Embedded Implementation}
\label{sec:embedded}

The estimator of Section~\ref{sec:estimator} is implemented directly on the microcontroller introduced in Section~\ref{sec:system_architecture}. In addition to executing the ESKF, the same core handles the IMU and barometer interfaces, receives and parses the radar detection stream, and transmits the resulting odometry to the flight controller. No companion computer is used in the estimation pipeline.

\subsection{Firmware Architecture and Scheduling}
\label{sec:firmware}

The firmware runs bare-metal on the $600\,\mathrm{MHz}$ Cortex-M7 of the Teensy 4.0 as a polled super-loop, without an RTOS or a DMA-driven sensor pipeline. Each iteration first services the available sensor interfaces, executes the corresponding estimator operations, and subsequently publishes an updated odometry estimate when required. This architecture keeps the complete sensor-to-estimate pipeline on a single processor core and avoids task scheduling and inter-process communication overhead.

The estimator operates asynchronously at the native rates of the three aiding sources described in Section~\ref{sec:system_architecture}. IMU measurements drive the prediction step at $200\,\mathrm{Hz}$. The sampling period is gated in software such that a faster execution of the main loop does not process the same accelerometer and gyroscope sample more than once. New barometric measurements trigger the scalar update of Section~\ref{sec:baro_factor} at $50\,\mathrm{Hz}$, while radar frames arrive at $25\,\mathrm{Hz}$. Once a radar frame has been received and parsed, its detections are processed sequentially using the radial-speed update of Section~\ref{sec:radar_factor}. The posterior after each accepted detection therefore becomes the prior for the subsequent detection in the frame.

Sensor acquisition and estimator timing use the microcontroller's \texttt{millis()} and \texttt{micros()} clocks rather than hardware-triggered sampling. Consequently, the implementation does not enforce deterministic sampling at the peripheral level, and some timing variation is introduced by the polled architecture. The radar frame period of $40\,\mathrm{ms}$ forms the principal deadline for the largest measurement update, while consecutive IMU predictions have a nominal separation of $5\,\mathrm{ms}$. The runtime measurements reported in Section~\ref{sec:results} are used to verify that the embedded implementation satisfies these timing requirements.

The ESKF is implemented using statically sized linear-algebra objects. The error state has fixed dimension $21$, the process-noise vector has dimension $12$, and each radar or barometric update uses a $1\times21$ measurement Jacobian. The maximum radar point-cloud size is also fixed at 1024 detections. These fixed dimensions allow the filter state, covariance, Jacobians, and intermediate matrices to be allocated statically, such that the estimator requires no dynamic memory allocation during operation.

All estimator arithmetic is performed in single precision. Linear algebra is provided by \texttt{ArduinoEigen}, a header-only port of \texttt{Eigen3} for embedded targets. The same estimator source, contained in \texttt{rio-lib}, is compiled both for the microcontroller and for the desktop replay used in Section~\ref{sec:results}. Both targets use the same single-precision scalar type and estimator equations. The replay therefore provides a direct implementation-parity reference rather than an independently implemented estimator.

The embedded and replay outputs are not expected to be numerically identical at every reported sample. On the microcontroller, the state is published following measurement events, whereas the desktop replay can expose the propagated state at the full $200\,\mathrm{Hz}$ IMU rate. Differences in sensor input handling and timing can therefore produce a residual even when the underlying estimator arithmetic is identical. The magnitude of this residual is evaluated separately in Section~\ref{sec:results}.

To characterise the computational cost without perturbing the estimator scheduling, the main processing stages are instrumented using the Cortex-M7 cycle counter. At the $600\,\mathrm{MHz}$ core clock, this provides sub-microsecond timing resolution. Separate measurements are collected for the IMU prediction, each scalar radar update, the complete radar-frame update, the barometric update, MAVLink transmission, and the complete main-loop iteration. For each stage, the firmware accumulates its execution count, running mean, and worst-case duration. This instrumentation is used together with the static memory footprint to quantify the computational requirements of the complete embedded pipeline in Section~\ref{sec:results}.

\subsection{Autopilot Interface}
\label{sec:autopilot_interface}

The embedded estimator is interfaced to an otherwise unmodified PX4 flight stack through its standard external-odometry interface. PX4 maps incoming MAVLink \texttt{ODOMETRY} messages to the \texttt{vehicle\_visual\_odometry} uORB topic, from which its internal EKF2 can fuse the externally estimated state \cite{px4_user_guide,mavlink_developer_guide}. This interface separates the proposed estimator from the flight-control stack: the radar-inertial state is computed entirely on the Teensy, while the Pixhawk receives it as an external aiding source for control.

An \texttt{ODOMETRY} message is transmitted following estimator measurement events, with the output rate limited to a nominal $50\,\mathrm{Hz}$. Each message contains the estimated position and orientation in \texttt{MAV\_FRAME\_LOCAL\_NED} and the estimated linear velocity in \texttt{MAV\_FRAME\_BODY\_FRD}. The corresponding pose and velocity covariance entries are obtained from the relevant blocks of the ESKF error-state covariance $\mathbf{P}$ and included with the message. This allows EKF2 to account for the uncertainty of the external estimate rather than receiving pose and velocity as noise-free quantities.

The internal estimator convention is converted to the NED and FRD conventions expected by PX4 only at the publication boundary. A constant rotation maps the estimator world and body frames to the corresponding PX4 frames, leaving the internal ESKF formulation of Section~\ref{sec:estimator} unchanged. The same transform includes a small $0.5^\circ$ pitch correction between the carrier-board IMU frame and the Pixhawk body frame, obtained by aligning the gravity direction of the published attitude with the attitude reported by EKF2. This ensures that the external attitude presented to PX4 is aligned with the airframe reference.

Which components of the external estimate EKF2 uses is controlled through the standard \texttt{EKF2\_EV\_CTRL} configuration. Position, velocity, and yaw aiding can therefore be enabled independently without modifying either PX4 or the embedded estimator. For the closed-loop flight reported in Section~\ref{sec:results}, horizontal and vertical position, three-dimensional velocity, and yaw from the proposed estimator are enabled. The native GPS, barometric, and magnetometer aiding of EKF2 are disabled, with external odometry selected as the height reference. PX4 consequently retains its own inertial propagation, but its exteroceptive state aiding is supplied by the embedded radar-inertial estimator.

This arrangement permits the same embedded estimate to be evaluated both independently and in closed-loop flight. In the former case, the published state is logged and compared against the references described in Section~\ref{sec:experimental_setup}. In the latter, EKF2 fuses the published odometry with the Pixhawk IMU and provides the state used by the standard PX4 position controller. No modification of the PX4 estimator or controller is required.

\section{Experimental Setup}
\label{sec:experimental_setup}

The complete sensor module of Section~\ref{sec:system_architecture} is mounted on a small multirotor together with a Pixhawk 6C Mini. The radar-inertial estimator runs entirely on the Cortex-M7 microcontroller. The GPS-aided PX4 EKF2 estimate is logged as an outdoor qualitative reference, while the Pixhawk is used as the flight controller in the closed-loop experiment described in Section~\ref{sec:autopilot_interface}.

\subsection{Environments and Datasets}
\label{sec:datasets}

The evaluation comprises five flight datasets spanning a forest, an open field, and an indoor motion-capture arena. The datasets and their corresponding references are summarised in Table~\ref{tab:datasets}.

\begin{table*}[t]
\centering
\caption{Flight datasets used in the evaluation.}
\label{tab:datasets}
\begin{tabular}{llll}
\toprule
\textbf{Dataset} & \textbf{Environment} & \textbf{Motion} &
\textbf{Reference} \\
\midrule
Forest (slalom)      & Forest                & Slalom between trees
& GPS-aided EKF2 (qualitative) \\
Field (fast)         & Open field            & Maximum-speed pass
& GPS-aided EKF2 (qualitative) \\
Indoor (square)      & Motion-capture arena  & Open-loop square pattern
& Motion capture \\
Indoor (circular)    & Motion-capture arena  & Open-loop circular pattern
& Motion capture \\
Indoor (closed-loop) & Motion-capture arena  & Closed-loop flight
& None \\
\bottomrule
\end{tabular}
\end{table*}

The indoor reference is provided by a Qualisys motion-capture system \cite{qualisys_miqus}. Its accuracy is substantially higher than the estimation errors considered here, and it is therefore treated as ground truth. Outdoors, the GPS-aided PX4 EKF2 estimate is used only as a qualitative reference. Its height and heading are not sufficiently reliable for quantitative evaluation, and short GPS dropouts occur during flight. No outdoor pose-error metrics are therefore reported.

The open-loop datasets are used to evaluate trajectory and velocity estimation without feeding the proposed estimate back to the flight controller. The indoor datasets additionally provide ground truth for quantitative pose-error evaluation. A desktop replay compiled from the same estimator source as the embedded implementation is used to evaluate microcontroller-to-desktop parity. Finally, the closed-loop dataset evaluates whether the embedded estimate is sufficiently consistent for position control through the standard PX4 external odometry interface.

To provide additional context for the evaluated motion, Table~\ref{tab:dataset_characteristics} summarises the duration, path length, and speed of the four open-loop sequences. The indoor values are obtained from the motion-capture reference whereas the outdoor values are obtained from the GPS-aided EKF2 reference.

\begin{table*}[t]
\centering
\caption{Duration, horizontal path length, mean speed, and peak
speed of the open-loop flight datasets over the evaluated intervals.}
\label{tab:dataset_characteristics}
\begin{tabular}{lrrrr}
\toprule
\textbf{Dataset} &
\textbf{Duration [s]} &
\textbf{Horizontal path [m]} &
\textbf{Mean speed [m/s]} &
\textbf{Peak speed [m/s]} \\
\midrule
Forest (slalom)    & 108 & 140 & 1.33 & 3.16 \\
Field (fast)       &  59 & 211 & 3.67 & 12.45 \\
Indoor (square)    & 145 &  43 & 0.38 & 1.56 \\
Indoor (circular)  &  99 &  73 & 0.78 & 1.86 \\
\bottomrule
\end{tabular}
\end{table*}

Unless stated otherwise, the radar innovation gate is set to $\sigma_{\mathrm{gate}}=3$, and both barometric aiding and the second-order underweighting of Section~\ref{sec:radar_factor} are enabled. None of the references used for evaluation are available to the proposed estimator, and all experiments use the radar configuration described in Section~\ref{sec:sensor_suite}.

\subsection{APE and RPE}
\label{sec:metrics}

Trajectory accuracy is evaluated using the absolute pose error (APE) and relative pose error (RPE) implemented in \cite{evo_traj}. Before evaluation, the estimated and reference trajectories are aligned at their origins. APE is reported for translation and measures the accumulated deviation from the ground-truth trajectory. RPE is evaluated over fixed $10\,\mathrm{m}$ trajectory segments and measures the local consistency of the estimate independently of the accumulated global drift.

Quantitative APE and RPE are reported only for the two open-loop indoor datasets, for which motion-capture ground truth is available. The outdoor datasets are evaluated qualitatively through trajectory and velocity agreement with the GPS-aided EKF2 reference.

\section{Results}
\label{sec:results}

\subsection{Open-Loop Estimation}
\label{sec:results_open_loop}

\begin{figure*}[t]
\centering
\includegraphics[width=0.32\textwidth]{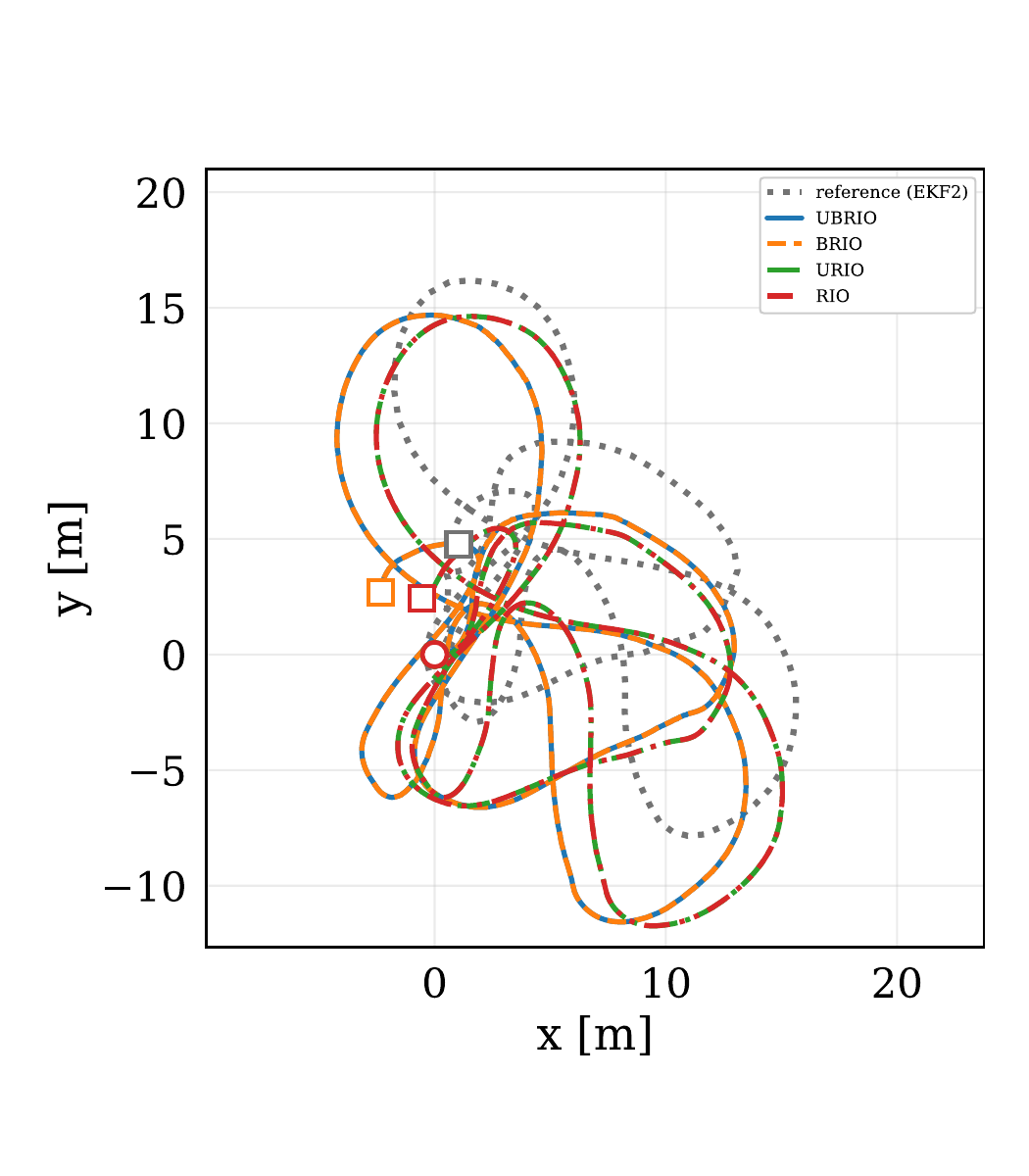}
\hfill
\includegraphics[width=0.32\textwidth]{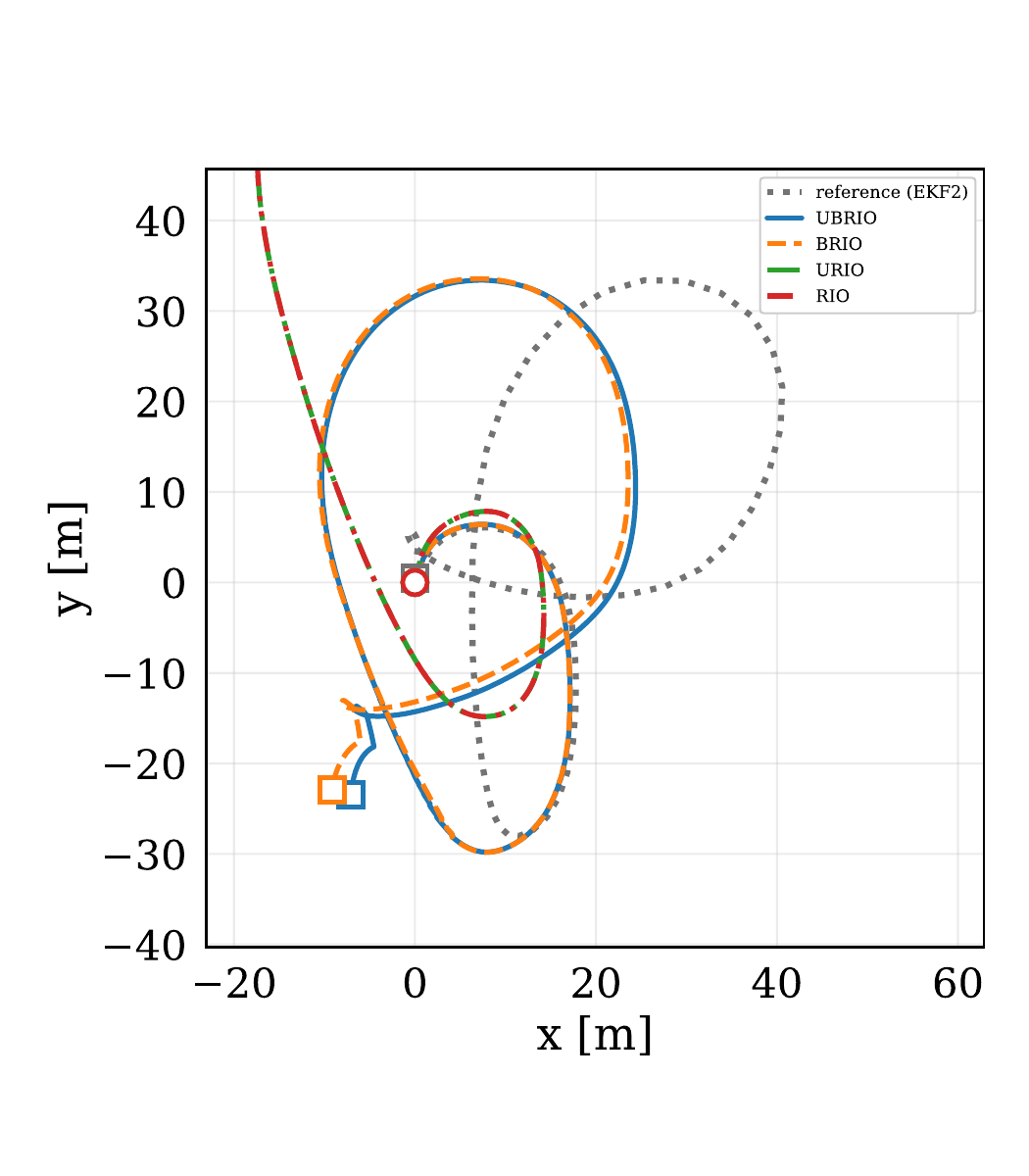}
\hfill
\includegraphics[width=0.32\textwidth]{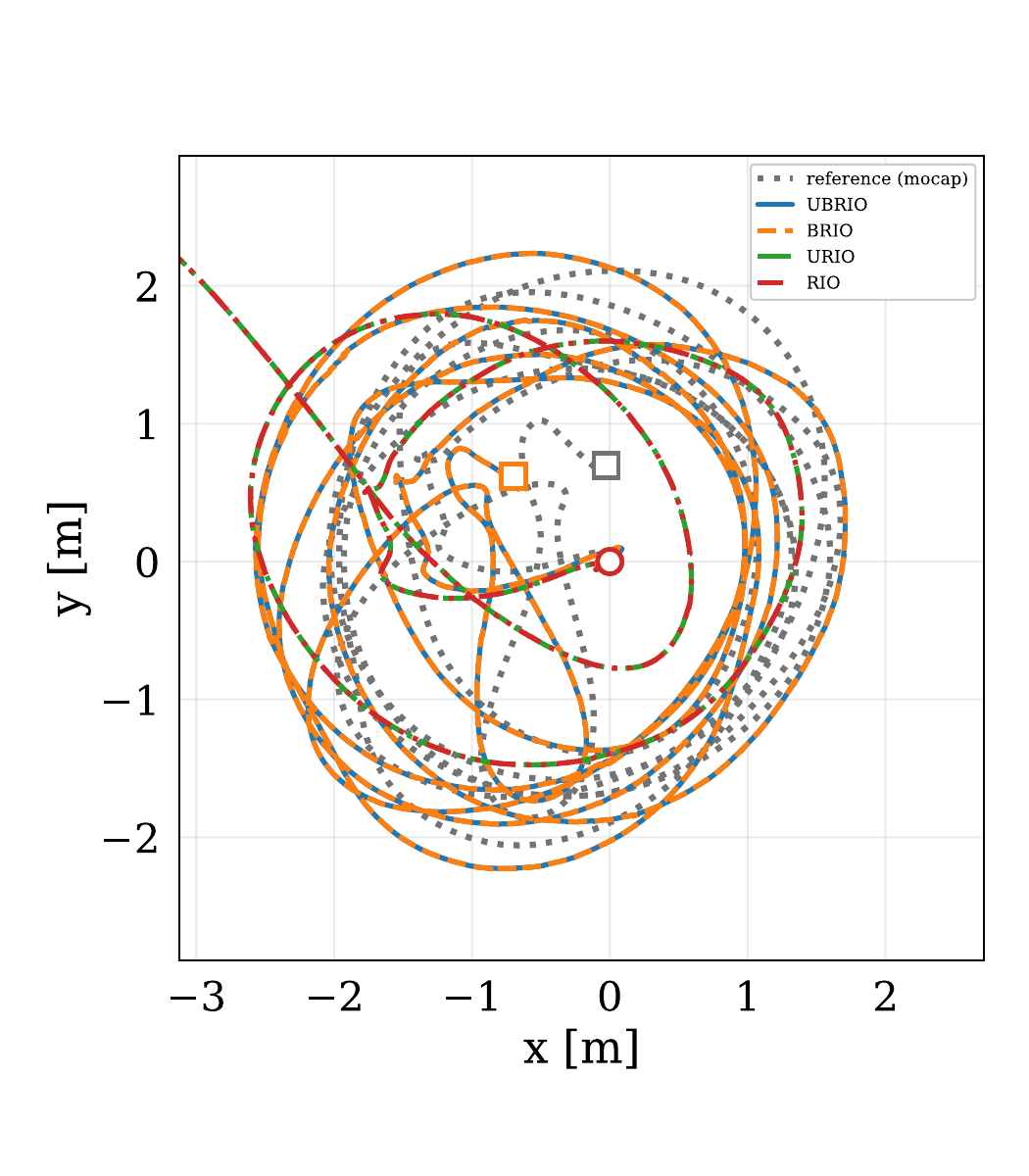}
\caption{Estimated horizontal trajectories against the available reference. From left to right: forest slalom, maximum-speed field pass, and indoor circular flight. Circles mark the start of each trajectory and squares mark the end. The outdoor reference is GPS-aided EKF2 and is used qualitatively, whereas the indoor reference is motion-capture ground truth.}
\label{fig:trajectories}
\end{figure*}

Figure~\ref{fig:trajectories} shows representative horizontal trajectories from the forest, high-speed field, and indoor circular flights. The estimator follows the overall trajectory in all three environments. Outdoors, the estimate agrees qualitatively with the GPS-aided EKF2 reference, while the indoor trajectory closely follows the motion-capture ground truth.

\begin{figure}[t]
\centering
\includegraphics[width=\columnwidth]{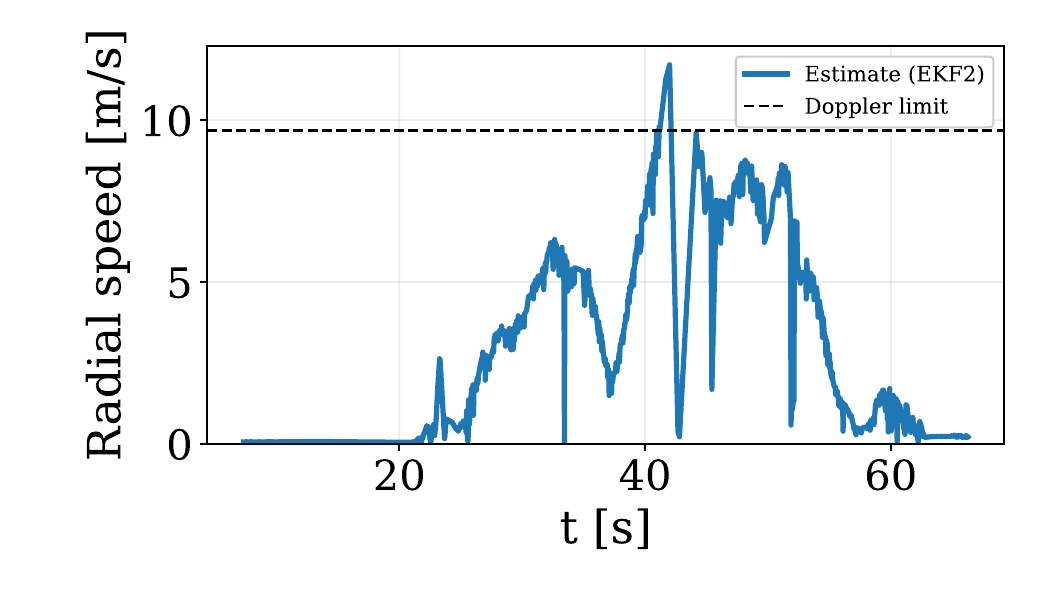}
\caption{The calculated doppler speed from EKF2 compared to the doppler limit for the field dataset. The platform's high velocity results in some few radar points exceeding this limit and thus experiencing aliasing.}
\label{fig:reference_doppler_field}
\end{figure}

\begin{figure*}[t]
\centering
\includegraphics[width=0.32\textwidth]{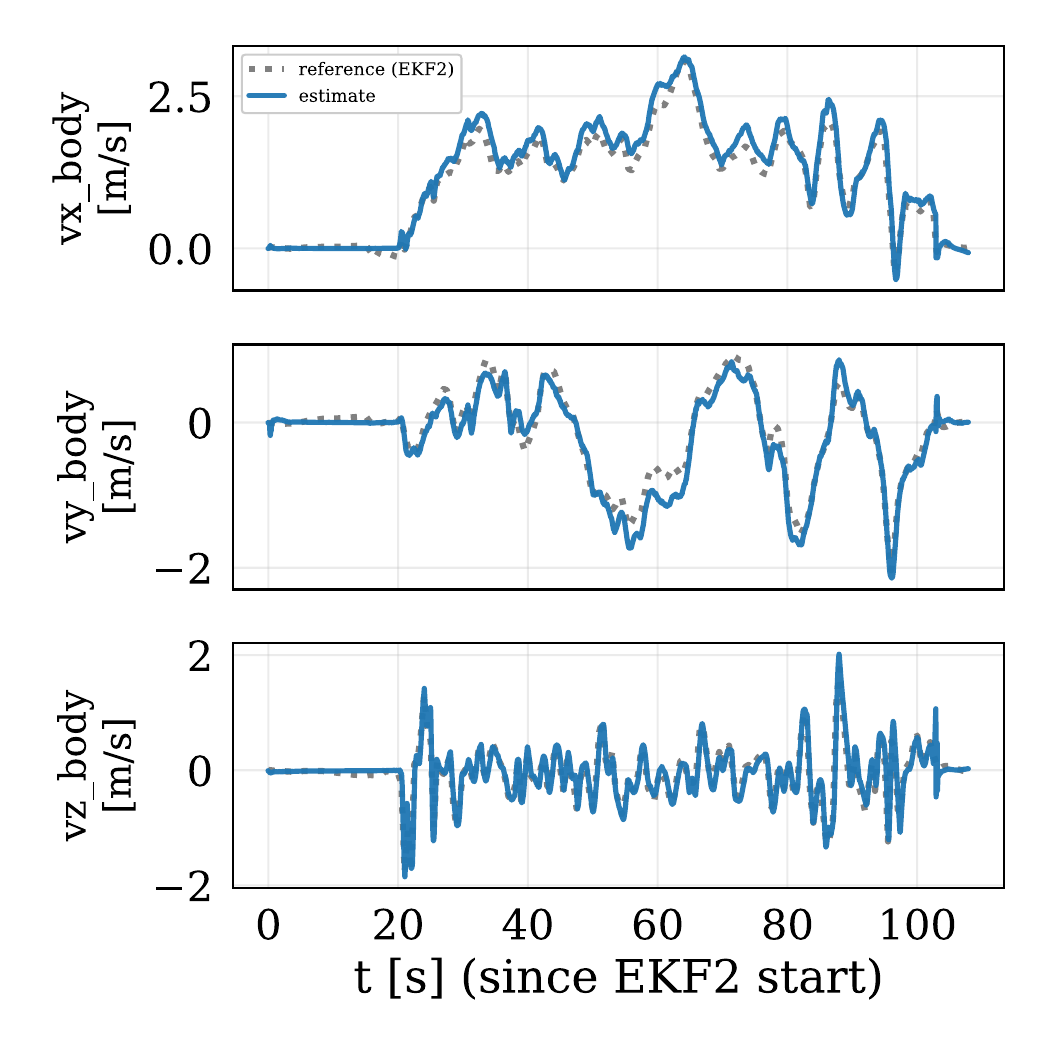}
\hfill
\includegraphics[width=0.32\textwidth]{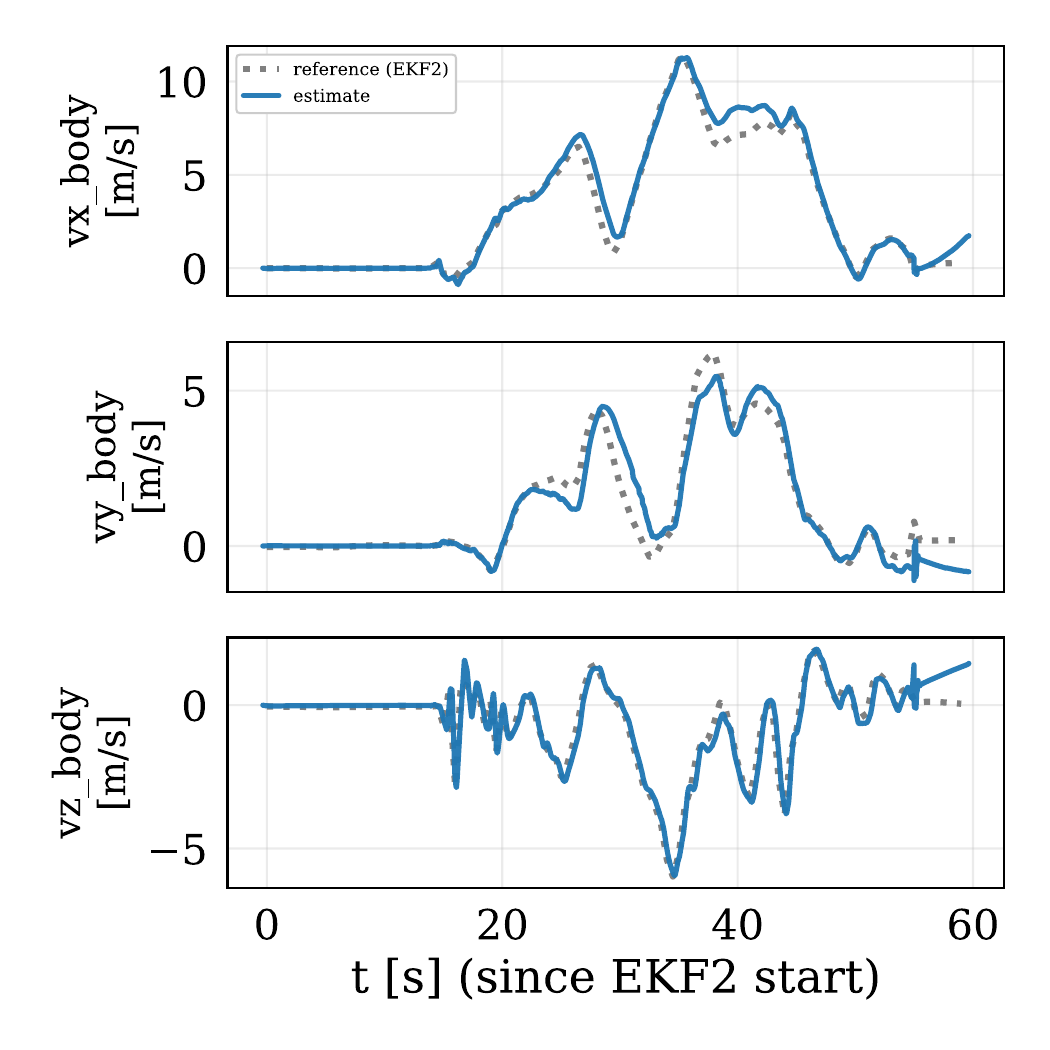}
\hfill
\includegraphics[width=0.32\textwidth]{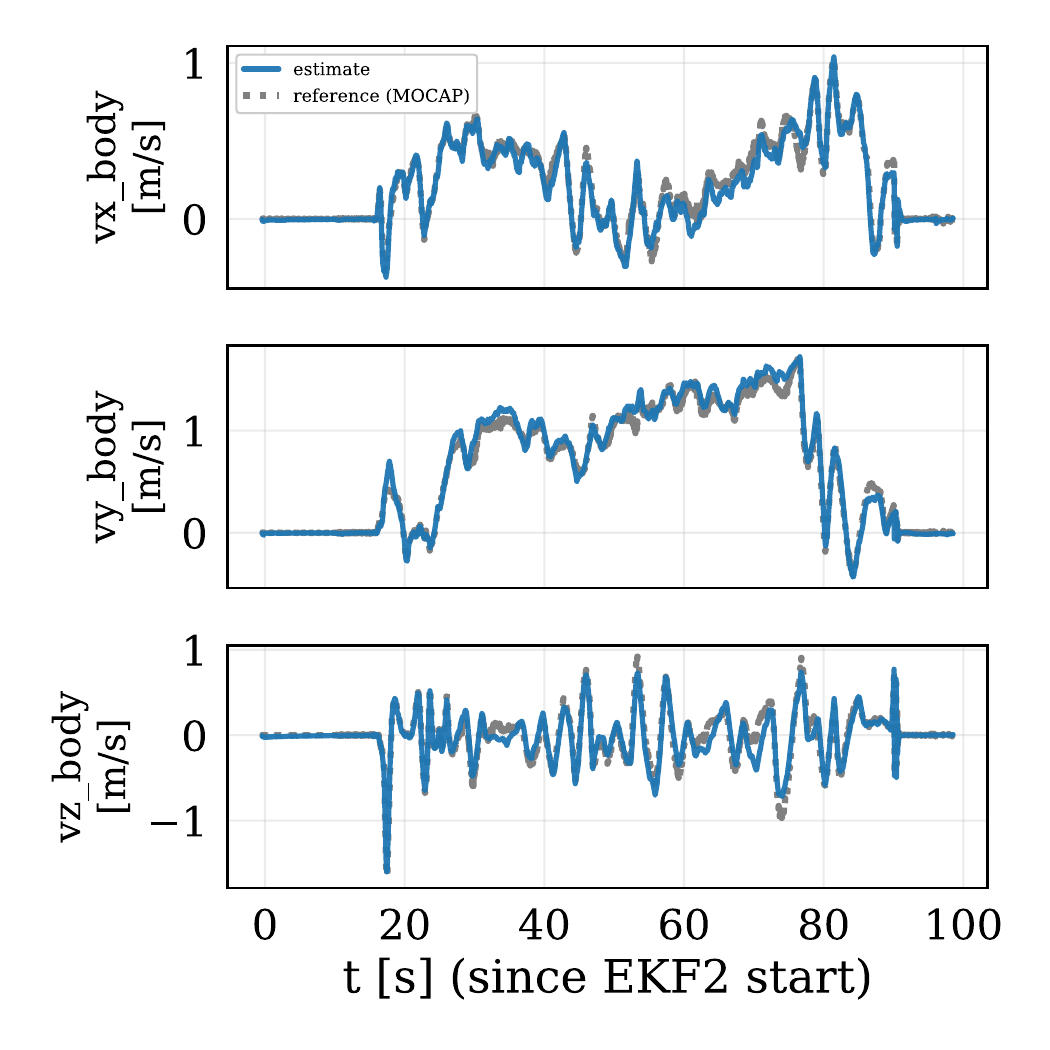}
\caption{Estimated velocity against the available reference. From left to right: forest slalom, maximum-speed field pass, and indoor circular flight. The outdoor reference is GPS-aided EKF2 and is used qualitatively, whereas the indoor reference is motion-capture ground truth.}
\label{fig:velocity}
\end{figure*}

The corresponding velocity estimates are shown in Figure~\ref{fig:velocity}. The radar-aided velocity follows the reference throughout the forest and indoor flights and remains consistent with the GPS-aided estimate during the maximum-speed pass. Indoors, the comparison is against motion-capture ground truth. The field dataset reaches a peak speed of $12.45\,\mathrm{m/s}$, exceeding the radar's $9.68\,\mathrm{m/s}$ maximum unambiguous radial speed, as seen in Figure~\ref{fig:reference_doppler_field}. Doppler aliasing is therefore expected during part of the sequence. The estimate does not diverge, although a residual between the proposed estimate and EKF2 becomes visible at the highest speed. 
The aliasing is clear from the predicted Doppler speed in Figure~\ref{fig:reference_doppler_field}, calculated from the radar velocity estimate of the EKF2 and the radar point bearing vectors. Although the radar sensor itself will not produce a Doppler measurement above the maximum \cite{nissov_characterization}, here we see several radar points with radial speed above the limit. These point are thus expected to return aliased measurements of Doppler therefore rendering them outliers. The estimator demonstrates that it is robust to brief periods of such phenomena.

\begin{figure}[t]
\centering
\includegraphics[width=\columnwidth]{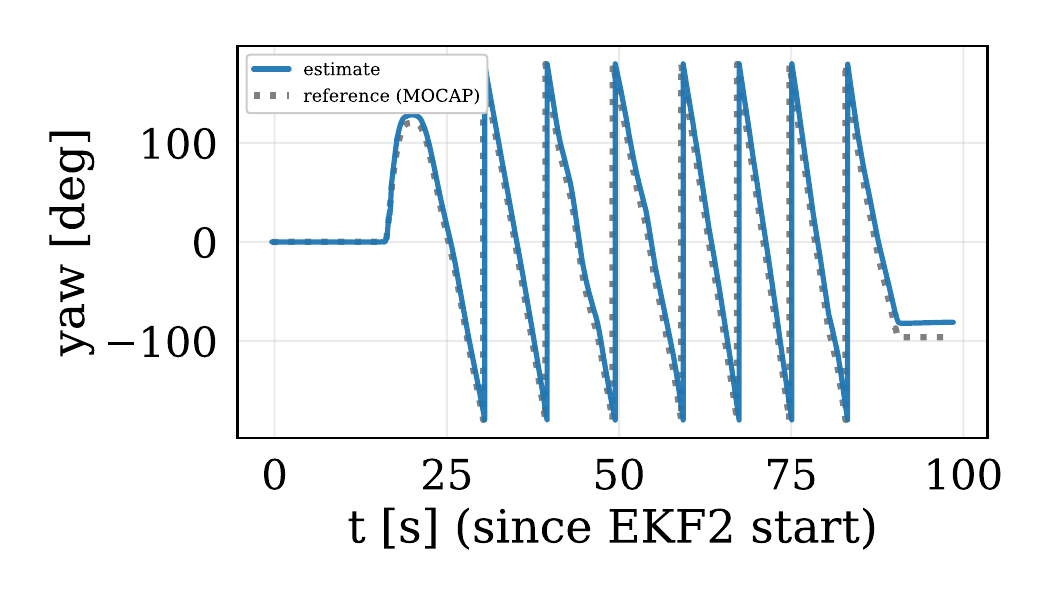}
\caption{Estimated yaw during the indoor circular flight against motion-capture ground truth.}
\label{fig:indoor_yaw}
\end{figure}

A slowly growing heading error is observed in the indoor flights and contributes to position drift, as shown for the indoor circular flight in Figure~\ref{fig:indoor_yaw}. The estimator has no absolute heading measurement, rendering yaw unobservable. Bias-induced heading drift is a separate issue since gyroscope-bias estimation depends on platform motion and coupling through inertial propagation, as well as on the direct lever-arm term in the radar measurement. The short radar--IMU lever arm reduces this observability.

Table~\ref{tab:pose_error} reports the translational APE and RPE for the two indoor flights. The square trajectory produces the larger APE. In contrast, the $10\,\mathrm{m}$ RPE remains similar for both flights, indicating comparable short-horizon estimation performance.

\begin{table}[t]
\centering
\caption{Translational APE and $10\,\mathrm{m}$ RPE against motion-capture ground truth. Values are RMSE $\pm$ one standard deviation.}
\label{tab:pose_error}
\begin{tabular}{lcc}
\toprule
\textbf{Dataset} &
\textbf{APE [m]} &
\textbf{RPE [m]} \\
\midrule
Indoor (square)   & $0.82 \pm 0.45$ & $0.28 \pm 0.09$ \\
Indoor (circular) & $0.51 \pm 0.29$ & $0.27 \pm 0.10$ \\
\bottomrule
\end{tabular}
\end{table}

Recent, comparable BRIO methods \cite{girod2024brio} report translational RPE RSME of approximately $2.9$--$5.0\,\%$ on the ColoRadar benchmark. Although the datasets differ, this provides a useful reference for the estimation performance obtained here on microcontroller-class hardware, which achieves comparable performance.

\subsection{Barometric Aiding and Second-Order Underweighting}
\label{sec:results_ablation}

For the following comparisons, UBRIO denoted radar-inertial odometry with barometric aiding and underweighting, BRIO denoted radar-inertial odometry with barometric aiding, URIO denoted radar-inertial odometry with underweighting, and RIO denoted radar-inetial odometry without either component.

The influence of the barometric factor and the second-order underweighting is evaluated by replaying the same measurements with all four combinations of the two components. The resulting indoor pose-error metrics are shown in Table~\ref{tab:ablation}.

\begin{table}[t]
\centering
\caption{Ablation of barometric aiding and second-order underweighting. RPE is evaluated over $10\,\mathrm{m}$ segments.}
\label{tab:ablation}
\begin{tabular}{llcc}
\toprule
\textbf{Dataset} & \textbf{Configuration}
& \textbf{APE [m]} & \textbf{RPE [m]} \\
\midrule
Circular
& Baro + underweighting & 0.513 & 0.266 \\
& Baro only              & 0.513 & 0.266 \\
& Radar + underweighting & 103.950 & 7.592 \\
& Radar only             & 103.952 & 7.592 \\
\midrule
Square
& Baro + underweighting & 0.821 & 0.279 \\
& Baro only              & 0.821 & 0.279 \\
& Radar + underweighting & 0.653 & 0.262 \\
& Radar only             & 0.653 & 0.262 \\
\bottomrule
\end{tabular}
\end{table}

\begin{figure}[t]
\centering
\includegraphics[width=\columnwidth]{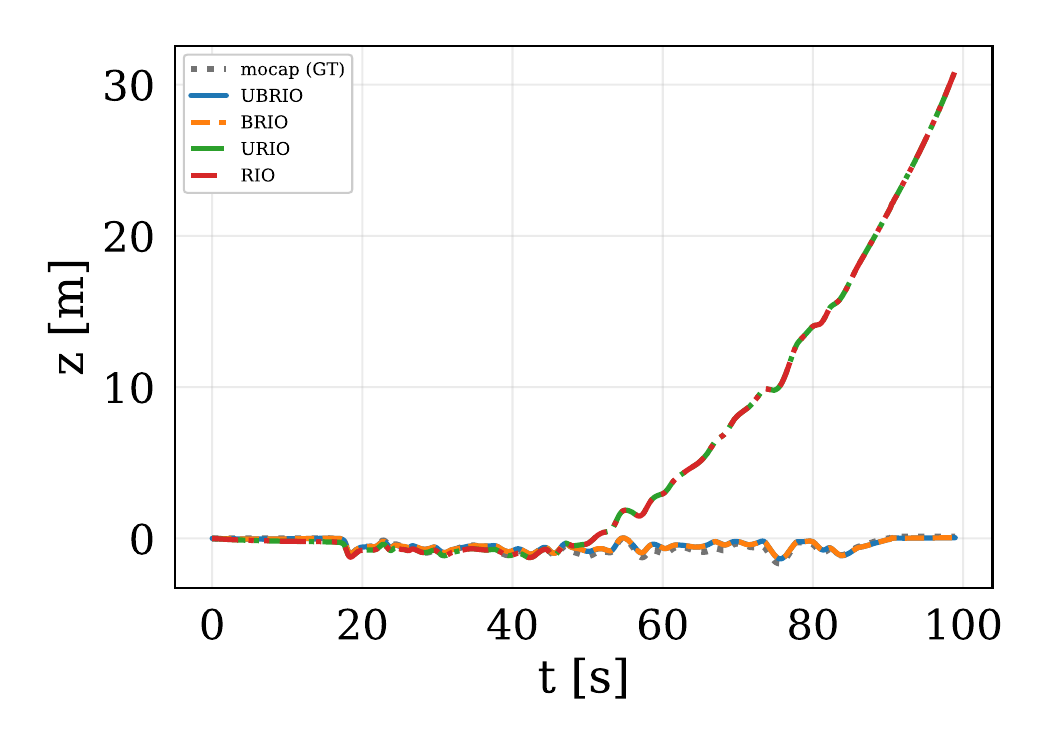}
\caption{Vertical position during the indoor circular flight for the four ablation configurations against motion-capture ground truth. The barometrically aided configurations do not diverge, whereas the radar-only configurations suffer from vertical drift.}
\label{fig:baro_altitude}
\end{figure}

The barometric measurement has a substantial effect on the vertical channel, as shown in Figure~\ref{fig:baro_altitude}. During the circular flight, the radar-only configurations develop large vertical drift and subsequently diverge, whereas both barometrically aided configurations remain bounded. The square flight has a shorter path and does not exhibit the same vertical divergence. Its radar-only APE is therefore slightly lower for this particular sequence.

Second-order underweighting has little effect on the indoor APE and RPE reported in Table~\ref{tab:ablation}. A small difference between the estimates with and without underweighting is visible during the high-dynamics field pass in Figure~\ref{fig:trajectories}. Since no reliable ground truth is available for this sequence, it is not possible to determine whether underweighting improves estimation accuracy.

\subsection{Embedded Parity and Resource Footprint}
\label{sec:results_embedded}

The embedded implementation is compared with a desktop replay compiled from the same single-precision estimator source. On the indoor circular flight, the translational difference between the embedded estimate and replay has an RMSE of $0.09\,\mathrm{m}$ and a maximum of $0.19\,\mathrm{m}$. On the square flight, the corresponding values are $0.05\,\mathrm{m}$ and $0.10\,\mathrm{m}$. The vertical RMSE is $0.04\,\mathrm{m}$ and $0.02\,\mathrm{m}$ for the circular and square flights, respectively.

These residuals are approximately $6$ to $16$ times smaller than the APE against motion capture in Table~\ref{tab:pose_error}. The remaining difference is partly explained by the different output rates. The embedded estimate is recovered from odometry publications following measurement events, whereas the replay exposes the propagated state at the full $200\,\mathrm{Hz}$ IMU rate. The remaining contribution was not isolated. Since both implementations compile the same estimator source using the same single-precision scalar type, the comparison indicates that the embedded arithmetic is not the dominant source of estimation error.

The hardware and memory footprint of the embedded implementation is summarised in Table~\ref{tab:resources}. The complete application, including the estimator, sensor drivers, linear-algebra library, MAVLink interface, and logging support, occupies $140.3\,\mathrm{KB}$ of flash. The fixed-size estimator performs no dynamic memory allocation during operation.

\begin{table}[t]
\centering
\caption{SWaP-C characteristics and static memory footprint of the embedded radar-inertial module.}
\label{tab:resources}
\begin{tabular}{ll}
\toprule
\textbf{Property} & \textbf{Value} \\
\midrule
Module footprint & $\approx 115 \times 25\,\mathrm{mm}$ \\
Mass & $\approx 50\,\mathrm{g}$ (estimated) \\
Power consumption & $\approx 3\,\mathrm{W}$ (estimated) \\
Prototype hardware cost & $\approx \$260$ \\
\midrule
Flash & $140.3\,\mathrm{KB}$ (6.9\%) \\
RAM1 (TCM) & $171.3\,\mathrm{KB}$ (33.5\%) \\
RAM2 (OCRAM) & $12.4\,\mathrm{KB}$ (2.4\%) \\
\bottomrule
\end{tabular}
\end{table}

The memory footprint leaves substantial margin on the present microcontroller. The execution times reported below are the binding constraint when moving to a less capable processor. The physical footprint of the module is primarily set by the IWR6843AOPEVM, which has approximately the same footprint as the custom carrier board. Further reductions in size, mass, power consumption, and cost would therefore require a more integrated radar implementation rather than reductions in estimator memory usage.

The measured execution times are reported in Table~\ref{tab:runtime}. The IMU prediction requires $233\,\mu\mathrm{s}$ on average, corresponding to less than $5\,\%$ of its $5\,\mathrm{ms}$ nominal sample period. With second-order underweighting enabled, a complete radar frame requires $6.44\,\mathrm{ms}$ on average and $13.69\,\mathrm{ms}$ in the worst observed case, remaining below the $40\,\mathrm{ms}$ radar frame period.

\begin{table}[t]
\centering
\caption{Execution time on the $600\,\mathrm{MHz}$ Cortex-M7, measured over the indoor circular sequence.}
\label{tab:runtime}
\begin{tabular}{lrr}
\toprule
\textbf{Stage} &
\textbf{Mean [$\mu$s]} &
\textbf{Worst [$\mu$s]} \\
\midrule
IMU prediction              & 233  & 242 \\
Radar update/point, UW      & 318  & 334 \\
Radar update/point, no UW   & 168  & 184 \\
Radar frame, UW             & 6443 & 13688 \\
Radar frame, no UW          & 3254 & 6963 \\
Barometer update            & 174  & 178 \\
MAVLink publish             & 2189 & 2271 \\
Main loop, UW               & 767  & 23861 \\
Main loop, no UW            & 677  & 17043 \\
\bottomrule
\end{tabular}
\end{table}

The numerical Hessian required by the underweighting approximately doubles the radar-update cost, from $168$ to $318\,\mu\mathrm{s}$ per detection and from $3.25$ to $6.44\,\mathrm{ms}$ per radar frame on average. Because radar frames are processed in only a fraction of the main-loop iterations, the mean complete-loop cost increases by about $13\,\%$.

The worst observed loop iteration with underweighting lasts $23.86\,\mathrm{ms}$. Although this remains below the radar-frame period, it spans several nominal IMU sample intervals. The present polled implementation reads only the most recent IMU sample, meaning that intermediate measurements can be missed during long iterations. The associated dynamics are therefore not represented explicitly in the prediction. This is a limitation of the firmware scheduling rather than the computational cost of the ESKF itself.

\subsection{Closed-Loop Flight}
\label{sec:results_closed_loop}

Finally, the estimator is evaluated in closed-loop flight through the PX4 interface of Section~\ref{sec:autopilot_interface}. Horizontal and vertical position, three-dimensional velocity, and yaw from the proposed estimator are enabled as external aiding in EKF2. Native GPS, barometric, and magnetometer aiding are disabled. PX4 retains its own IMU propagation and fuses the radar-inertial estimate as an external measurement.

\begin{figure}[t]
\centering
\includegraphics[width=\columnwidth]{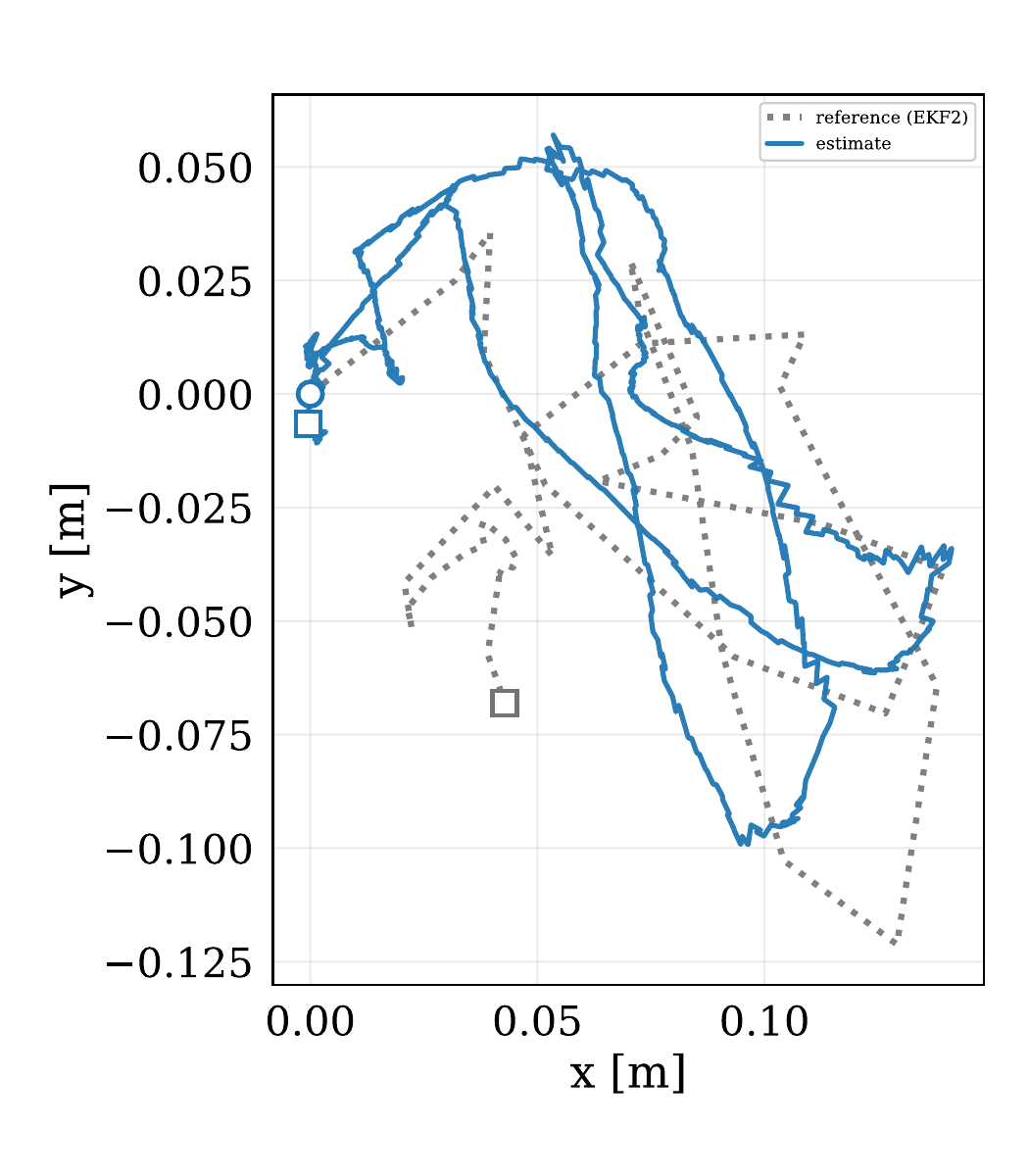}
\caption{Estimated horizontal trajectory during the closed-loop
flight. Circles mark the start of each trajectory and squares mark the end. No independent ground truth is available for this sequence.}
\label{fig:closed_loop}
\end{figure}

The platform remains controllable and holds position under manual guidance using this configuration. Figure~\ref{fig:closed_loop} shows the estimated horizontal trajectory. No independent ground truth is available for this flight, and the result is therefore interpreted as a control demonstration rather than a quantitative accuracy test.

The heading and vertical-drift limitations observed in the open-loop experiments are also apparent during closed-loop flight. Heading drift requires corrective yaw input from the pilot, while a residual vertical velocity error results in a slow change in altitude. Barometric aiding limits vertical position drift but does not eliminate errors in the estimated vertical velocity. Since no independent ground truth is available for this flight, these effects are not quantified. The closed-loop result is therefore used as a demonstration that the embedded estimate can be used for position control through an unmodified PX4 flight stack.

\section{Conclusion}
\label{sec:conclusion}

This work demonstrates a tightly coupled radar-inertial estimator that executes together with its sensor drivers and measurement handling on a single $600\,\mathrm{MHz}$ Cortex-M7 microcontroller. The estimator incorporates individual radar radial-speed measurements, barometric altitude aiding, and online radar--IMU extrinsic estimation, while the complete embedded implementation uses fixed-size single-precision linear algebra and no dynamic memory allocation.

Flight experiments in forest, open-field, and indoor environments show that the estimator tracks both platform trajectory and velocity. In the motion-capture arena, the two open-loop trajectories achieve translational APE RMSEs of $0.82\,\mathrm{m}$ and $0.51\,\mathrm{m}$, with corresponding $10\,\mathrm{m}$ RPE RMSEs of $0.28\,\mathrm{m}$ and $0.27\,\mathrm{m}$. Outdoors, where a sufficiently accurate ground truth is unavailable, the estimated trajectory and velocity agree qualitatively with the GPS-aided PX4 reference.

The anchored differential barometric factor is important to the vertical channel and prevents the large accumulated altitude drift observed in the radar-only ablation. The second-order underweighting has little measurable influence on the indoor accuracy metrics, while approximately doubling the computational cost of each radar update. A visible difference appears only during the most aggressive flight sequence, for which no quantitative ground truth is available.

The implementation remains well within the available memory of the microcontroller, and each individual estimator stage executes within its corresponding sensor period. 
Together, these results show that tightly coupled radar-inertial odometry of this form is feasible on microcontroller-class hardware without a companion computer.

Closed-loop flight further demonstrates that the estimate can be provided through the documented PX4 external-odometry interface and used for position control without modifying the flight stack. The dominant remaining estimation limitation is yaw drift. With no absolute heading measurement and only a short radar--IMU lever arm, the gyroscope yaw-rate bias remains weakly observable. In addition, the current polled firmware can miss IMU samples during long radar-processing iterations. Future development should therefore prioritise additional heading aiding and buffered, interrupt-driven sensor acquisition. A controlled evaluation in degraded visibility would additionally test the environmental resilience that motivates the use of radar.

\acknowledgements %
This material was supported by European Commission under Horizon Europe Grants SYNERGISE (EC 101121321) and AUTOASSESS (101120732).

\bibliographystyle{IEEEtran}
\bibliography{references}

\thebiography
\begin{biographywithpic}
{Nicolai Adil Øyen Aatif}{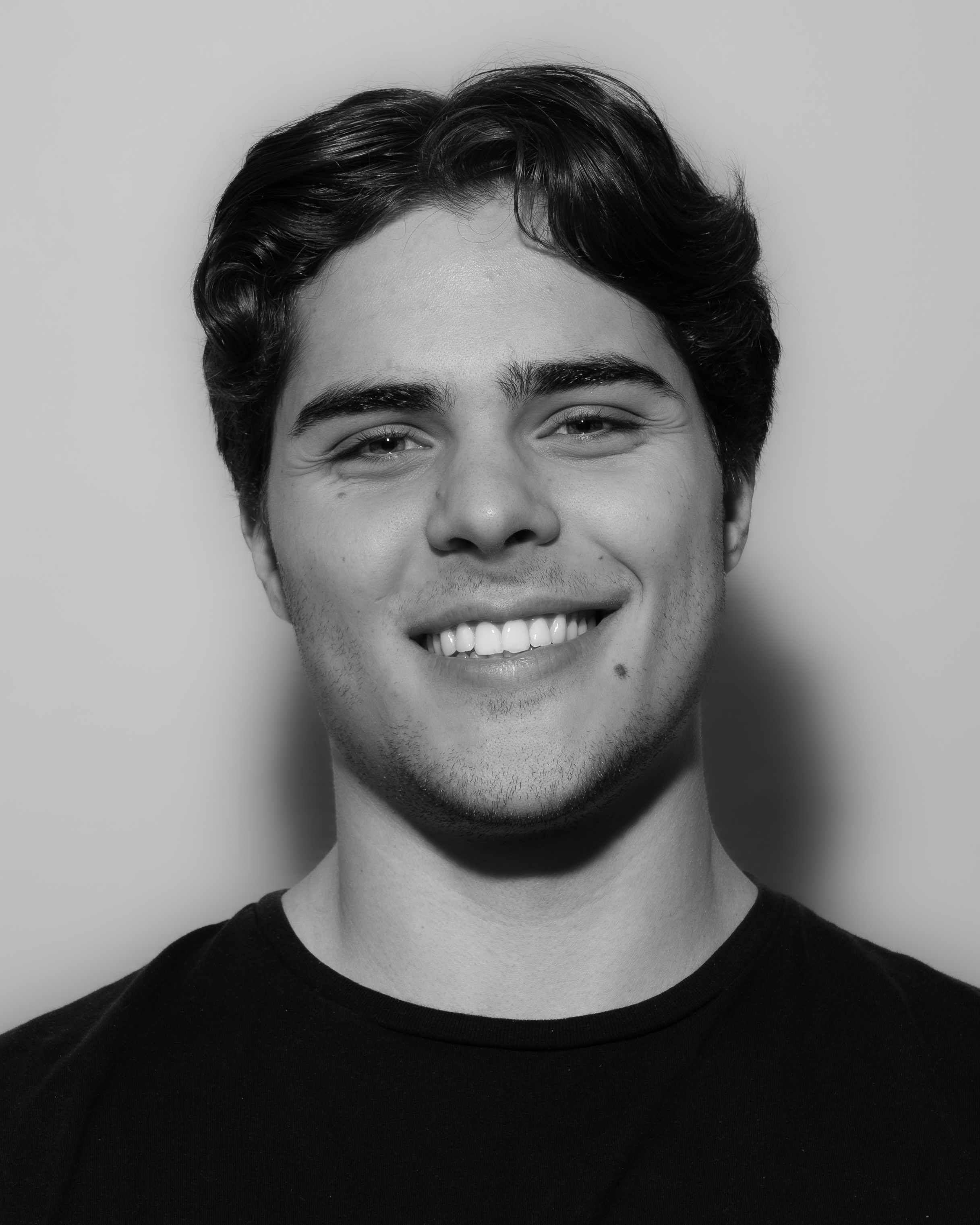}
has a background in Cybernetics and Robotics from the Norwegian University of Science and Technology (NTNU), specialising in autonomous systems. He currently works at Six Robotics in Oslo as part of the perception team. His work and interests include computer vision, sensor fusion, navigation, and autonomous robotic systems.
\end{biographywithpic} 

\begin{biographywithpic}
{Morten Nissov}{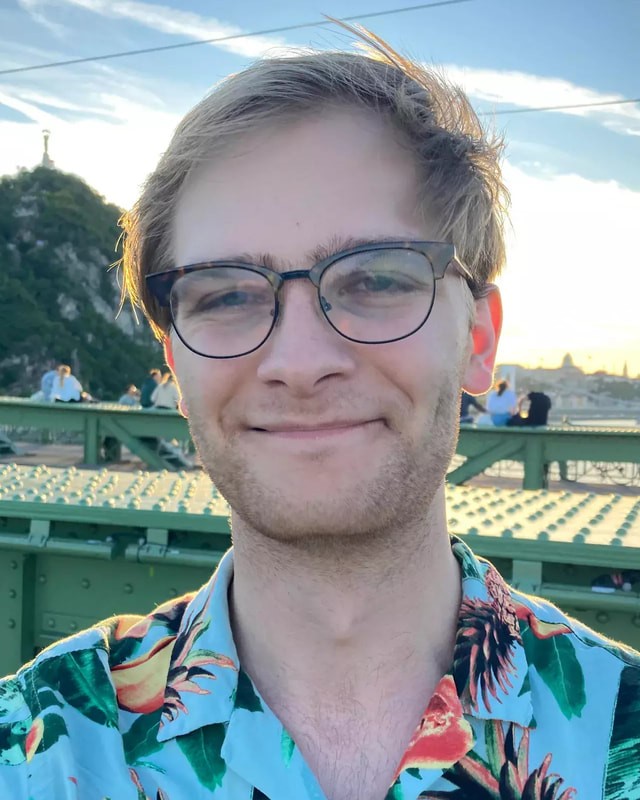}
received the M.Sc. degree in electrical engineering from the Technical University of Denmark (DTU) in 2021, and the Ph.D. degree in engineering cybernetics from the Norwegian University of Science and Technology (NTNU) in 2026. He is currently working on developing robust estimation solutions for autonomous robot systems at FieldAI. His research interests include robot perception, estimation, and control, with a focus on the development of algorithms exploiting multi-sensor information in degraded conditions.

\end{biographywithpic}

\begin{biographywithpic}
{Kostas Alexis}{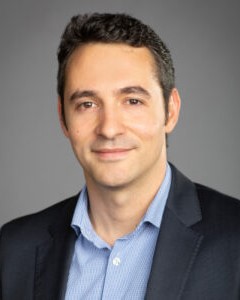}
is a Professor at the Department of Engineering Cybernetics at the Norwegian University of Science and Technology (NTNU), head of the Autonomous Robot Lab, and Director of the Norwegian Centre for Embodied AI. Together with his team, he conducts research on resilient robotic autonomy, exploring how autonomous systems can operate in high-risk, uncertain environments by presenting resourcefulness, robustness, and redundancy. Focusing on enhancing and safeguarding the autonomy capabilities of robotic systems, his research cuts across model-based optimization for control, sensor fusion, path planning, and learning algorithms for navigation. Prof. Alexis has served as Principal Investigator in major international grants both in Europe and the US, and was the PI and team lead of Team CERBERUS, winners of the DARPA Subterranean Challenge.
\end{biographywithpic}

\end{document}